\documentclass[journal]{IEEEtran}
\usepackage{amsmath}
\usepackage{amssymb}
\usepackage{graphicx}
\usepackage{caption}
\usepackage{color}
\usepackage{midfloat}
\usepackage{float}

\ifCLASSINFOpdf

\else

\fi

\newcommand{\xdownarrow}[1]{%
  {\left\downarrow\vbox to #1{}\right.\kern-\nulldelimiterspace}
}

\begin{document}

\title{Discrete Cosserat Approach for Multi-Section Soft Robots Dynamics}

\author{Federico~Renda$^{1*}$,~\IEEEmembership{Member,~IEEE,},
		Fr\'ed\'eric Boyer$^{2}$,~\IEEEmembership{Member,~IEEE,},
		Jorge~Dias$^{1}$,~\IEEEmembership{Member,~IEEE,} and 
		Lakmal~Seneviratne$^{1}$,~\IEEEmembership{Member,~IEEE,}
\thanks{* Corresponding author {\tt\small federico.renda@kustar.ac.ae}}
\thanks{$^{1}$ F. Renda, J. Dias and L. Seneviratne are with the Khalifa University Robotics Institute, Khalifa University, Abu Dhabi, UAE.}
\thanks{$^{2}$ F. Boyer is with the IRCCyN, Ecole de Mines de Nantes, Nantes, France.}
}

\markboth{Transaction on Robotics,~Vol.~, No.~, June~2016}%
{Renda \MakeLowercase{\textit{et al.}}: Discrete Cosserat Approach for Multi-Section Soft Robots Dynamics}

\maketitle

\begin{abstract}

In spite of recent progress, soft robotics still suffers from a lack of unified modeling framework. Nowadays, the most adopted model for the design and control of soft robots is the piece-wise constant curvature model, with its consolidated benefits and drawbacks. In this work, an alternative model for multi-section soft robots dynamics is presented based on a discrete Cosserat approach, which, not only takes into account shear and torsional deformations, essentials to cope with out-of-plane external loads, but also inherits the geometrical and mechanical properties of the continuous Cosserat model, making it the natural soft robotics counterpart of the traditional rigid robotics dynamics model. The soundness of the model is demonstrated through extensive simulation and experimental results for both plane and out-of-plane motions.

\end{abstract}

\section{Introduction}

Since the beginning of the soft robotics field, many researchers have contributed in the development of mathematical modeling approaches which could be able to describe the kinematics and dynamics of such infinite Degrees of Freedom (DoF) robots, while addressing the challenging requirements imposed by their robotic applications \cite{Rus2015}, \cite{Kim_TB2013}. In order to meet the standards achieved in traditional rigid robotics, a model for soft robotics should be at the same time computational inexpensive and sufficiently accurate. Furthermore, it should be able to shed light on the mathematical sub-models and to encompass them in a unified framework. Such a modeling framework is the necessary condition for developing the physical designs and control architectures of these new soft robots as well as their task-related motions and path planning.

Despite the short history of soft robotics, important results have been already achieved and several complementary modeling approaches have been proposed to date. Those approaches can be divided into three main categories: Piece-wise Constant Curvature (PCC) models, continuum Cosserat models and 3D Finite Elements Models (FEM).

The PCC modeling approach is by far the most adopted in the soft robotics community \cite{Webster_IJRR2010}. It represents the soft robot as a finite collection of circular arcs, which can be described by only three parameters (radius of curvature, angle of the arc and bending plane), a simplification which drastically reduces the number of variables needed. Originally devoted to kinematics modeling \cite{Jones_TRO2006}, this approach has been extended and improved over the years with excellent results as in \cite{Godage_IJRR2015,Falkenhahn_TRO2015}. In spite of this success, the constant curvature assumption is not always valid, especially when the robot is subject to non-negligible external loads including gravity.

The continuum Cosserat approach is an infinite DoF model where the soft robot is represented by continuously stacking an infinite number of infinitesimal micro-solids. It has been primarily used in the context of hyper-redundant robot \cite{Chirikjian_IROS1993}, and more recently applied to soft robotics locomotion \cite{Boyer_2006,Boyer_BB_2015} and manipulation \cite{Rucker2011}, in both static \cite{Renda_BB2012} and dynamic \cite{Renda_TRO} conditions. This approach has been also extended to shell-like soft robots for underwater locomotion inspired by cephalopods \cite{Renda_BB2015}, \cite{Renda_ISRR2015}. Despite their accuracy and fidelity to the continuous mechanics of the soft robots, the resulting partial differential equations are computationally demanding and difficult to use for control purposes.

Finally, a FEM based approach has also been explored for modeling and real-time control of soft robots \cite{Largilliere_ICRA2015}. It is so far limited to quasi-static conditions, and needs linearisation of the structural elasticity that may not apply to many soft robot geometries.

Although it might be impossible, due to physical reasons, to achieve the same elevate standard reached by the mathematical models for rigid robotics, the research outlined above constitute a significant attempt in this direction. In the present paper, we build upon the two main pillars achieved so far to obtain, in the authors opinion, one of the most promising approach towards a unified mathematical framework between traditional and soft robotics. Going further into details, the continuous model developed with the Cosserat approach is discretized in order to implement the PCC idea of reducing the dimension of the configuration space by assuming a piece-wise constant deformation along the soft manipulator. As a consequence, the soft manipulator is completely described by a finite set of strain vectors which plays the same role as that of the joint vector for traditional robotics.

The strains allowed by the Cosserat approach include torsion and shears along with curvature and elongation. Thus, we call this method Piece-wise Constant Strain (PCS) model. With respect to the PCC model, the PCS model not only takes into account shears and torsion, which are both essential to cope with out-of-plane external loads, but also shares a common geometric structure with the equations of motion of their rigid robotics counterpart. As a matter of fact, the PCS model provides a direct forward kinematics between the joint space and the task space without any intermediate map. Furthermore, based on the $ SE(3) $ geometry of the Cosserat approach, it guarantees a closer relation with the rigid body geometry of the traditional robotics. Finally, the discrete Cosserat framework allows the adaptation of different actuation solutions and external loads models, including the interaction with a dense medium, without any significant changes in the structure of the model, and is so more independent from the specific applications.

Part of the present work has been presented in the conference paper \cite{Renda_IROS2016}. Beyond this work, the full multi-section dynamics is addressed here and a recursive algorithm for calculating the coefficients of the dynamics equations is presented. Furthermore, the homogeneity with the standard rigid robotics theory is highlighted and extensive simulations along with experimental results are shown for the multi-section dynamics case. In the following, in section \ref{CCM} the continuous Cosserat model is briefly reminded in order to introduce the discretization developed in the subsequent section \ref{DCM}. Finally, the model is corroborated through extensive simulation and experimental results in sections \ref{SR} and \ref{ER} for the general case of a soft manipulator operating in a dense medium like water.

\section{Continuous Cosserat Model}\label{CCM}

In the Cosserat theory, the configuration of a micro-solid of a soft body with respect to the inertial frame at a certain time is characterized by a position vector $ u $ and an orientation matrix $ R $, parameterized by the material abscissa $ X \in [0, L] $ along the robot arm. Thus, the configuration space is defined as a curve $ g(\cdot): X \mapsto g(X) \in SE(3) $ with
$$ g = \left( \begin{array}{cc} R & u \\ 0 & 1 \end{array} \right) .$$

Then, the strain state of the soft arm is defined by the vector field along the curve $ g(\cdot) $ given by $ X \mapsto \widehat{\xi}(X)=g^{-1}\partial g/\partial X=g^{-1}g' \in \mathfrak{se}(3) $, where the hat is the isomorphism between the twist vector representation and the matrix representation of the Lie algebra $ \mathfrak{se}(3) $. The components of this field are specified in the (micro-)body frames as: 
$$ \widehat{\xi} = \left( \begin{array}{cc} \tilde{\mathsf{k}} & \mathsf{q} \\ 0 & 0 \end{array} \right) \in \mathfrak{se}(3) \; \text{,} \; \xi=\left( \mathsf{k}^T , \mathsf{q}^T \right)^T \in \mathbb{R}^{6} \; \text{,} $$
where $ \mathsf{q}(X) $ represents the linear strains, and $ \mathsf{k}(X) $ the angular strains. The tilde is the isomorphism between three dimensional vectors and skew symmetric matrices.

The time evolution of the configuration curve $ g(\cdot) $ is represented by the twist vector field $ X \mapsto \eta(X) \in \mathbb{R}^{6} $ defined by $ \widehat{\eta}(X)=g^{-1}\partial g/\partial t=g^{-1}\dot{g} $. This field can be detailed in terms of their components in the (micro-)body frames as: 
$$ \widehat{\eta} = \left( \begin{array}{cc} \tilde{\mathsf{w}} & \mathsf{v} \\ 0 & 0 \end{array} \right) \in \mathfrak{se}(3) \; \text{,} \; \eta=\left(  \mathsf{w}^T , \mathsf{v}^T \right)^T \in \mathbb{R}^{6} \; . $$
Where $ \mathsf{v}(X) $ and $ \mathsf{w}(X) $ are respectively the linear and angular velocity at a given time instant.

\subsection{Continuous Kinematics} \label{CK}
Given the above construction, we can obtain the kinematic equations relating the strains of the robot arm $ \xi $ with the position $ g $, velocity $ \eta $ and acceleration $ \dot{\eta} $ for each infinitesimal micro-solid constituting the robot.
By definition, the first equation is given by:
\begin{equation}\label{kinem_pos}
g' = g\widehat{\xi} \; .
\end{equation}

Then, the equality of mixed partial derivatives $ (\dot{g})'=\dot{(g')} $ gives the following compatibility equation between strain and velocity:
\begin{equation}\label{kinem_vel}
\eta' = \dot{\xi} - \mathrm{ad}_{\xi}\eta \; \text{,}
\end{equation}
where $ \mathrm{ad} $ is the adjoint map defined as (together with the coadjoint map $ \mathrm{ad}^* $): 
$$ \mathrm{ad}_{\xi}=\left( \begin{array}{cc} \widetilde{\mathsf{k}} & 0 \\ \widetilde{\mathsf{q}} & \widetilde{\mathsf{k}} \end{array} \right) \; \text{,} \; \mathrm{ad}_{\xi}^*=-\mathrm{ad}_{\xi}^{T}=\left( \begin{array}{cc} \widetilde{\mathsf{k}} & \widetilde{\mathsf{q}} \\ 0 & \widetilde{\mathsf{k}} \end{array} \right) \; . $$

Finally, by taking the derivative of (\ref{kinem_vel}) with respect to time, we obtain the continuous model of acceleration:
\begin{equation}\label{kinem_acc}
\dot{\eta}' = \ddot{\xi} - \mathrm{ad}_{\dot{\xi}}\eta - \mathrm{ad}_{\xi}\dot{\eta} \; .
\end{equation}

\subsection{Continuous Dynamics}
In \cite{Boyer_JNLS_2016} it is shown that Cosserat beam dynamics can be directly derived from the extension to continuum media of a variational calculus originally introduced by H. Poincar\'e \cite{Poincare_1901}. In contrast to usual Lagrangian mechanics, this calculus allows deriving the dynamics of a system whose the configuration space definition requires the structure of Lie group. In this context, the dynamics of the Cosserat medium can be entirely derived from a Lagrangian density $ \mathfrak{T}(\eta)-\mathfrak{U}(\xi) $, where $ \mathfrak{T} $ and $ \mathfrak{U} $ are functions of the Lie algebra vectors modelling the densities of kinetic and elastic energy of the Cosserat beam per unit of material length $ X $. Applying this variational calculus to such a density leads to the strong form
of a Cosserat beam with respect to the micro-solid frames.
\begin{equation}\label{dynamics}
\mathcal{M} \dot{\eta} + \mathrm{ad}^*_{\eta}\left(\mathcal{M}\eta\right) = \mathcal{F}_{i}'+\mathrm{ad}^*_{\xi}\mathcal{F}_{i} + \bar{\mathcal{F}}_{a} + \bar{\mathcal{F}}_{e} \; \text{,}
\end{equation}
where $ \mathcal{F}_{i}(X)=\partial \mathfrak{U}/\partial X $ is the wrench of internal forces, $ \bar{\mathcal{F}}_{a}(X,t) $ is the distributed actuation loads, $ \bar{\mathcal{F}}_{e}(X) $ is the external wrench of distributed applied forces and $ \mathcal{M}(X) $ is the screw inertia matrix. Let us specify the angular and linear components of the internal and external wrenches: $ \mathcal{F}_{i}=\left( \mathsf{M}_{i}^T, \mathsf{N}_{i}^T \right)^T $, $ \bar{\mathcal{F}}_{a}=\left( \mathsf{m}_{a}^T, \mathsf{n}_{a}^T \right)^T $, $ \bar{\mathcal{F}}_{e}=\left( \mathsf{m}_{e}^T, \mathsf{n}_{e}^T \right)^T \in \mathbb{R}^6 $, where $ \mathsf{N}_{i}(X) $ and $ \mathsf{M}_{i}(X) $ are the internal force and torque vectors, $ \mathsf{n}_{a}(X,t) $ and $ \mathsf{m}_{a}(X,t) $ are the actuation force and torque inputs, while $ \mathsf{n}_{e}(X) $ and $ \mathsf{m}_{e}(X) $ are the external force and torque for unit of $ X $. By choosing a local (micro-)body frame oriented as in figure \ref{kinematics}, with the $ X $ axis pointing toward the tip of the robot arm and the $ Y $ and $ Z $ axes laying on the plane of the section (considered symmetric), the screw inertia matrix is equal to: $ \mathcal{M} =diag(J_x, J_y, J_z, A, A, A)\rho $, where $ \rho $ is the body density, $ A $ is the section area and $ J_y $, $ J_z $, $ J_x $ are respectively the bending and torsion second moment of inertia of the beam cross-section.

\begin{figure}
\centering
\includegraphics[scale=0.35]{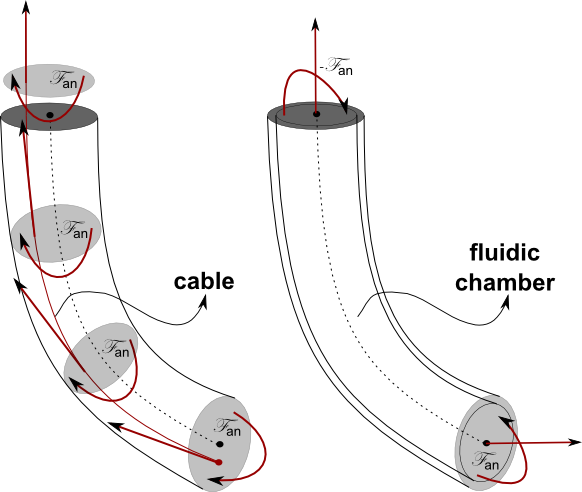}
\caption{Schematic of the cable-driven and fluidic actuation for one section.}
\label{actuation}
\end{figure}

Let us now specify the models of the distributed actuation, external load and internal forces appearing in (\ref{dynamics}) for the general case of a soft robot arm moving in a dense surrounding medium like water. Considering the two most important actuation systems implemented in soft robotics, the cable driven and the fluidic actuation \cite{Rus2015}, we have respectively: 
\begin{equation}\label{act_wrench}
\bar{\mathcal{F}}_{a}(X,t) = -\left(\mathcal{F}_{a}'+\mathrm{ad}_{\xi}^{*}\mathcal{F}_{a} \right) \; \text{and} \; \bar{\mathcal{F}}_{a}(X,t) = 0 \; \text{,}
\end{equation}
where $ \mathcal{F}_{a} $ is the cable wrench acting on the micro-solid  given by the cable tension and the cable path from the tip to the base \cite{Renda_TRO}, \cite{Rucker2011}, \cite{Renda_JMR2017}. The model of the fluidic actuator, widely used in soft robotics nowadays \cite{Polygerinos2015}, condensates the action of the pressure in a concentrated load at the tip of the section (Fig. \ref{actuation}).

Regarding the wrench of internal passive forces, a linear visco-elastic constitutive model, based on the Kelvin Voigt assumptions, is chosen \cite{Renda_TRO}.
\begin{equation}\label{int_wrench}
\mathcal{F}_{i}(X) = \Sigma \left( \xi - \xi^{0} \right) + \Upsilon \dot{\xi} \text{,}
\end{equation}
where $ \Sigma $  and $ \Upsilon $ are constant screw stiffness and viscosity matrices, equal to $ \Sigma = diag(GJ_x, EJ_y, EJ_z, EA, GA, GA) $, $ \Upsilon = diag(J_x, 3J_y, 3J_z, 3A, A, A)\upsilon $, $ E $ being the Young modulus, $ G $ the shear modulus and $ \upsilon $ the shear viscosity modulus; $\xi^{0}=[0 \; 0 \; 0 \; 1 \; 0 \; 0]^{T}$ stands for the zeros strain vector in the reference straight configuration. No other assumptions except the constitutive model are needed to describe the elastic behavior of the robot arm. 

As for the external loads, we have considered the general case of underwater operation, {\it i.e.} distributed loads due to gravity and buoyancy, drag, added mass and a concentrated/point load due to externally applied loads or contacts \cite{Renda_TRO}:
\begin{equation}\label{ext_wrench}
\begin{array}{l}
\bar{\mathcal{F}}_{e} = \left(1-\rho_w/\rho \right)\mathcal{M}\mathrm{Ad}_{g_rg(X)}^{-1}\mathcal{G} - \mathcal{D}||\mathsf{v}|| \eta + \delta(X-\bar{X})\mathcal{F}_p \; \text{,} \\
\mathcal{M}_a = \mathcal{M}+\mathcal{A}
\end{array}
\end{equation}
where $ \rho_w $ is the water density, $ \mathcal{G}=[0 \; 0 \; 0 \; -9.81 \; 0 \; 0]^{T} $ is the gravity twist with respect to the inertial frame (in accordance with the choice of inertial frame given in figure \ref{kinematics}), $ g_r $ is the transformation between the spatial frame and the base frame of the soft manipulator, $ \mathcal{D}(X) $ is the drag fluid dynamics coefficient, $ \delta(\cdot) $ is the Dirac distribution, $ \mathcal{F}_p $ is the wrench corresponding to the point load applied at $ \bar{X} $ and $ \mathcal{A}(X) $ is the added mass fluid dynamics coefficient. Note here that replacing $ \mathcal{M} $ by $ \mathcal{M}_a $ in (\ref{dynamics}) allows modeling inertial hydrodynamics forces exerted along the arm. Finally, we have introduced the Adjoint representation ($ \mathrm{Ad} $) of Lie group $ SE(3) $, defined as (together with the coAdjoint map $ \mathrm{Ad}^* $):
$$ \mathrm{Ad}_{g} = \left( \begin{array}{cc} R & 0 \\ \widetilde{\mathsf{u}}R & R \end{array} \right) \; \text{,} \; \mathrm{Ad}_{g}^*=\mathrm{Ad}_{g}^{-T}=\left( \begin{array}{cc} R & \widetilde{\mathsf{u}}R \\ 0 & R \end{array} \right) \; . $$
Finally, when the soft arm is working in a sparse medium like air we will let $ \rho_w $, and consequently $ \mathcal{D} $ and $ \mathcal{A} $, be equal to zero.

\section{Discrete Cosserat Model}\label{DCM}

Equations (\ref{kinem_pos}), (\ref{kinem_vel}), (\ref{kinem_acc}) and (\ref{dynamics}) of the continuous Cosserat model are suitable to model the kinematics and dynamics of soft robots expressing a non-constant deformation, as it has been presented in \cite{Renda_TRO} (and \cite{Renda_BB2015}, \cite{Renda_ISRR2015} for bi-dimensional bodies). In the subsequent development, we unify the constant and non-constant cases under the same mathematical framework. To that end, the continuous model is discretized by an analytic spatial integration. This is allowed by the piece-wise constant strain assumption which provides the condition to analytically integrate the continuum model and leads to the extension of the piece-wise constant curvature model, by including torsion and shears, without any additional effort. Furthermore, a profound and useful parallelism with the rigid manipulators theory can be achieved, which leads to the soft robot counterpart of the Lagrangian model of rigid serial manipulators.

\subsection{Piece-wise Constant Strain Kinematics}
At any instant $ t $, considering the strain field $ \xi(X) $ constant along each of the $ N $ sections of the soft arm, we can replace the continuous field with a finite set of $ N $ twist vectors $ \xi_{n} $ ($ n \in \{1,2,..,N\} $), which play the role of the joint vectors of traditional rigid robotics. Under this assumption, equation (\ref{kinem_pos}) becomes an homogeneous, linear, matrix differential equation with constant coefficients, which can be analytically solved at any section $ n $ using the matrix exponential method with the appropriate interval of $ X $ and initial value \cite{Edwards}. Going further into details, the material abscissa $ X \in [0,L] $ is divided into $ N $ sections of the form $ [0, L_1) $, $ (L_1, L_2) \dots (L_{N-1}, L_{N}] $ (with $ L_N = L $) and the initial value for the differential equation of the section $ n $ is given by the solution at the right boundary of the previous section ($ X = L_{n-1} $). In other words, the solutions are glued together, one on top of the other. With these considerations, the integration of (\ref{kinem_pos}) at a certain instant $ t $ becomes:
\begin{equation}\label{int_pos}
g(X) = g(L_{n-1})e^{\left( X-L_{n-1} \right) \widehat{\xi}_n} \; .
\end{equation}
It turns out that the infinite series of the exponential in (\ref{int_pos}) can be expressed in a compact way as follows \cite{Selig}:
\begin{equation}\label{exp_rototra}
\begin{split}
e^{\left( X-L_{n-1} \right) \widehat{\xi}_n} = I_4 + \left( X-L_{n-1} \right)\widehat{\xi}_n& \\
+ \frac{1}{\theta_n^2}\left(1-\cos\left(\left(X-L_{n-1}\right)\theta_n\right)\right)\widehat{\xi}_n^2& \\
+ \frac{1}{\theta_n^3}\left(\left(X-L_{n-1}\right)\theta_n-\sin\left(\left(X-L_{n-1}\right)\theta_n\right)\right)\widehat{\xi}_n^3 &=: g_n(X)\; \text{,}
\end{split}
\end{equation}
where $ \theta_n^2 = \mathsf{k_n}^T\mathsf{k_n} $. For straight configurations of the section, we have $ \widehat{\xi}_n^2 = 0_4 $ and hence: 
$$ e^{\left( X-L_{n-1} \right) \widehat{\xi}_n} = I_4 + \left( X-L_{n-1} \right)\widehat{\xi}_n \; \text{,} $$
which allows circumventing the well known singularity of straight arm pose of the PCC models \cite{Tatlicioglu_IROS2007}, \cite{Rone_TRO2014}. Equation (\ref{exp_rototra}) can be viewed as the $ SE(3) $ counterpart of the Rodrigues formula in $ SO(3) $. Calling $ g_n(X) $ the exponential function in (\ref{exp_rototra}), 
equation (\ref{int_pos}) can be written in the more familiar way:
\begin{equation}\label{int_pos2}
g(X) = g(L_{n-1})g_n(X) \; \text{,}
\end{equation}
which recursively returns the position and orientation of the micro-solid at $ X $ knowing the set of strains $ \xi_n $ only.

Similarly, the velocity of each micro-solid $ \eta(X) $ can be obtained by a piece-wise integration of the continuum model (\ref{kinem_vel}). Under constant strains condition, at each section $ n $ and time $ t $, equation (\ref{kinem_vel}) is a non-homogeneous, linear, matrix differential equation with constant coefficients (reminds that also $ \dot{\xi}_n $ is piece-wise constant) which can be analytically solved using the variation of parameters method, with the appropriate initial value \cite{Edwards}.
\begin{equation}\label{int_vel}
\begin{split}
\eta(X) = &e^{-\left( X-L_{n-1} \right) \mathrm{ad}_{\xi_n}} \\
&\left( \eta(L_{n-1}) + \int_{L_{n-1}}^{X} e^{\left( s-L_{n-1} \right) \mathrm{ad}_{\xi_n}} ds \dot{\xi}_n \right)\; .
\end{split}
\end{equation}
Again, the exponential function in (\ref{int_vel}) can be expressed with a finite number of terms \cite{Selig} (for the sake of presentation, $ x = X-L_{n-1} $ holds in the following).
\begin{equation}\label{exp_Adj}
\begin{split}
e^{x\mathrm{ad}_{\xi_n}} = I_6 + \frac{1}{2\theta_n}(3\sin(x\theta_n)-x\theta_n\cos(x\theta_n))\mathrm{ad}_{\xi_n}& \\
+ \frac{1}{2\theta_n^2}(4-4\cos(x\theta_n)-x\theta_n\sin(x\theta_n))\mathrm{ad}_{\xi_n}^2& \\
+ \frac{1}{2\theta_n^3}(\sin(x\theta_n)-x\theta_n\cos(x\theta_n))\mathrm{ad}_{\xi_n}^3& \\
+ \frac{1}{2\theta_n^4}(2-2\cos(x\theta_n)-x\theta_n\sin(x\theta_n))\mathrm{ad}_{\xi_n}^4& \\
= \mathrm{Ad}_{g_n(X)}&\; ,
\end{split}
\end{equation}
where for straight configurations we have $ \mathrm{ad}_{\xi_n}^2= 0_6 $ and thus, taking the limit for $ \theta_n\rightarrow 0 $, $ e^{x\mathrm{ad}_{\xi_n}} = I_6 + x\mathrm{ad}_{\xi_n} $. Thanks to the fact that $ e^{\mathrm{ad}_{\xi}}=\mathrm{Ad}_{e^{\widehat{\xi}}} $ (\cite{Abate} pg. 403 Lemma 7.5.9), we can notice that the exponential function (\ref{exp_Adj}) is nothing else but the Adjoint representation of the Lie group transformation $ g_n(X) $ of (\ref{exp_rototra}).
With this definition at hand, equation (\ref{int_vel}) can be rewritten as follows:
\begin{equation}\label{int_vel2}
\eta(X) = \mathrm{Ad}_{g_n(X)}^{-1} \left( \eta(L_{n-1}) + \mathrm{AD}_{g_n}(X) \dot{\xi}_n \right) \; \text{,}
\end{equation}
where we have defined:
\begin{equation}\label{int_coef}
\begin{split}
\mathrm{AD}_{g_n}(X) := \int_{L_{n-1}}^{X}\mathrm{Ad}_{g_n(s)}ds = \\
xI_6 + \frac{1}{2\theta_n^2}(4-4\cos(x\theta_n)-x\theta_n\sin(x\theta_n))\mathrm{ad}_{\xi_n} \\
+ \frac{1}{2\theta_n^3}(4x\theta_n-5\sin(x\theta_n)+x\theta_n\cos(x\theta_n))\mathrm{ad}_{\xi_n}^2 \\
+ \frac{1}{2\theta_n^4}(2-2\cos(x\theta_n)-x\theta_n\sin(x\theta_n))\mathrm{ad}_{\xi_n}^3 \\
+ \frac{1}{2\theta_n^5}(2x\theta_n-3\sin(x\theta_n)+x\theta_n\cos(x\theta_n))\mathrm{ad}_{\xi_n}^4 \; .
\end{split}
\end{equation}
Remarkably, equation (\ref{int_vel2}) recursively compute the velocity of any micro-solid at $ X $ along the soft arm as a function of the set of strains $ \xi_n $ and strain rates $ \dot{\xi} _n $.

Finally, the acceleration of any micro-solid at $ X $ ($ \dot{\eta}(X) $) can be calculated at any time $ t $, by means of a piece-wise integration of the continuous equation (\ref{kinem_acc}). Considering constant strains along one section, equation (\ref{kinem_acc}) is a non-homogeneous, linear, matrix differential equation with non-constant coefficients (given by the term $ \mathrm{ad}_{\dot{\xi}_n}\eta $ which is not constant with respect to $ X $ due to $ \eta(X) $). A direct application of the variation of parameters method with the appropriate initial value gives:
\begin{equation}\label{int_acc}
\begin{split}
\dot{\eta}(X) = &e^{-x \mathrm{ad}_{\xi_n}} \\
&\left( \dot{\eta}(L_{n-1}) + \int_{L_{n-1}}^{X} e^{x\mathrm{ad}_{\xi_n}}\left( \ddot{\xi}_n-\mathrm{ad}_{\dot{\xi}_n}\eta \right) ds \right) \; .
\end{split}
\end{equation}
Then, by virtue of the definitions of $ \mathrm{Ad}_{g_n} $, $ \mathrm{AD}_{g_n} $, we obtain:
\begin{equation}\label{int_acc2}
\begin{split}
&\dot{\eta}(X) = \mathrm{Ad}_{g_n(X)}^{-1} \\
&\left( \dot{\eta}(L_{n-1}) + \mathrm{AD}_{g_n}(X) \ddot{\xi}_n - \int_{L_{n-1}}^{X}\mathrm{Ad}_{g_n(s)}\mathrm{ad}_{\dot{\xi}_n}\eta(s) ds \right) \; \text{.}
\end{split}
\end{equation}
Let us focus on the term $ \mathrm{Ad}_{g_n(s)}\mathrm{ad}_{\dot{\xi}_n}\eta(s) $ inside the integral of the right end side. First, by means of equation (\ref{int_vel2}) and the properties of the adjoint map, we can write: 
$$ \mathrm{Ad}_{g_n(s)}\mathrm{ad}_{\dot{\xi}_n}\eta(s) = \mathrm{ad}_{\mathrm{Ad}_{g_n(s)}\dot{\xi}_n}\left( \eta\left( L_{n-1} \right)+\mathrm{AD}_{g_n}(s)\dot{\xi}_n \right) \; . $$
Then, evoking the linearity and anticommutativity of the adjoint map, and using equations (\ref{exp_Adj}) and (\ref{int_coef}), we obtain the equivalence
$$ \mathrm{Ad}_{g_n(s)}\mathrm{ad}_{\dot{\xi}_n}\eta(s) = \mathrm{ad}_{\mathrm{Ad}_{g_n(s)}\dot{\xi}_n}\eta\left( L_{n-1} \right) \; \textit{,} $$
which once substituted in (\ref{int_acc2}) (and using twice the anticommutativity of the adjoint map to make appear $ \mathrm{AD}_{g_n} $), gives the model of accelerations as follows:
\begin{equation}\label{int_acc3}
\dot{\eta}(X) = \mathrm{Ad}_{g_n}^{-1}\left( \dot{\eta}(L_{n-1}) - \mathrm{ad}_{\mathrm{AD}_{g_n}\dot{\xi}_n}\eta\left( L_{n-1} \right) + \mathrm{AD}_{g_n} \ddot{\xi}_n \right).
\end{equation}
Again, equation (\ref{int_acc3}) returns the acceleration of any micro-solid at $ X $ by means of the set of strains $ \xi_n $, strain rates $ \dot{\xi}_n $ and rates of strain rate $ \ddot{\xi}_n $ only.

\begin{figure}
\centering
\includegraphics[scale=0.25]{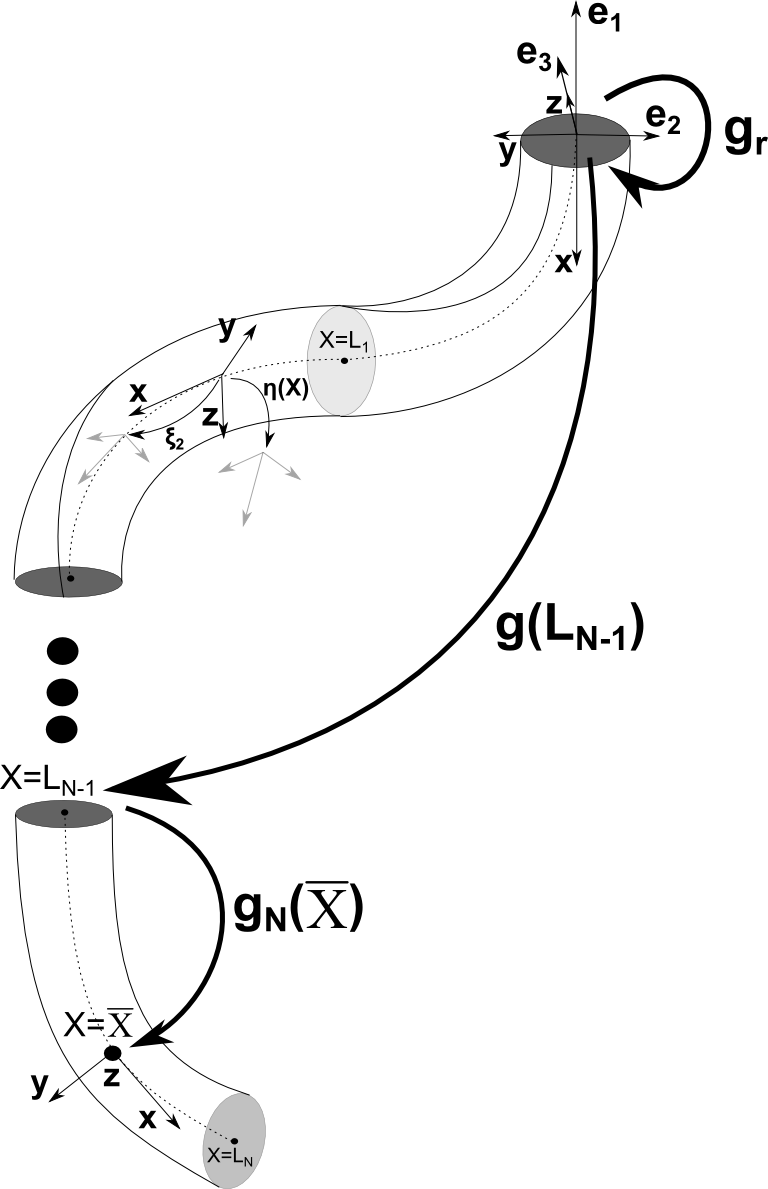}
\caption{Schematic of the kinematics of the piece-wise constant strain model.}
\label{kinematics}
\end{figure}

In order to develop the discrete Cosserat dynamic model for soft robots a relation between the kinematics quantities $ \eta $, $ \dot{\eta} $ and a joint vector for soft robotics needs to be established. To do so, we back track to the base the velocity term $ \eta(L_{n-1}) $ on the right end side of (\ref{int_vel2}), which becomes:
\begin{equation}\label{int_vel3}
\eta(X) = \sum_{i=1}^{n} \left( \prod_{j=n}^{i} \mathrm{Ad}_{g_j\left(min \left( L_j,X \right)\right)}^{-1} \right) \mathrm{AD}_{g_i}\left( min \left( L_i,X \right) \right) \dot{\xi}_i \; \text{.}
\end{equation}
where $ j $ is a descending index, we have considered a fixed base ($ \eta(0)=0_{6 \times 1} $) and $ L_{n-1}<X \leqslant L_{n} $. Introducing the soft robots joint vector
$$ \overrightarrow{\xi}=\left[ \xi_1^T \; \xi_2^T \cdots \; \xi_N^T \right]^T \in \mathbb{R}^{6N} \; \text{,} $$
equation (\ref{int_vel3}) can be expressed as:
\begin{equation}\label{int_vel4}
\eta(X) =  J(X)\dot{\overrightarrow{\xi}} \; \text{,}
\end{equation}
which define the softs robot geometric Jacobian $ J(X) \in \mathbb{R}^{6\times6N} $, shown in (\ref{Jacobian}).
\begin{figure*}[tb]
\begin{equation}\label{Jacobian}
J(X)=
\left\{ \begin{array}{lcl} \left[ \left(\mathrm{Ad}_{g_1}^{-1}\mathrm{AD}_{g_1}\right)(X) \;\; 0_6 \;\; \cdots \right] & \mbox{if} & 0<X \leqslant L_1 \\
\\
\left[ \mathrm{Ad}_{g_2(X)}^{-1} \left( \mathrm{Ad}_{g_1}^{-1}\mathrm{AD}_{g_1} \right) (L_1) \;\; \left( \mathrm{Ad}_{g_2}^{-1}\mathrm{AD}_{g_2} \right)(X) \;\; 0_6 \;\; \cdots \right] & \mbox{if} & L_1<X \leqslant L_2 \\
\\
\left[ \mathrm{Ad}_{g_3(X)}^{-1} \mathrm{Ad}_{g_2(L_2)}^{-1} \left( \mathrm{Ad}_{g_1}^{-1}\mathrm{AD}_{g_1} \right) (L_1) \;\; \mathrm{Ad}_{g_3(X)}^{-1} \left( \mathrm{Ad}_{g_2}^{-1}\mathrm{AD}_{g_2} \right)(L_2) \;\; \cdots \;\; 0_6 \right] & \mbox{if} & L_1<X \leqslant L_2 \\
\\
\vdots & \vdots & \vdots \\
\\
\left[ \prod\limits_{1}^{j=N} \mathrm{Ad}_{g_j\left(min \left( L_j,X \right)\right)}^{-1} \mathrm{AD}_{g_1}(L_1) \;\; \cdots \;\; \left( \mathrm{Ad}_{g_N}^{-1} \mathrm{AD}_{g_N} \right)(X) \right] & \mbox{if} & L_{N-1}<X \leqslant L_N
\end{array}\right.
\end{equation}
\end{figure*}

It is important to notice that the Jacobian (\ref{Jacobian}) is calculated directly from the strains $ \xi_n $ by means of equations (\ref{exp_Adj}) and (\ref{int_coef}). For this reason, in accordance with the rigid manipulators theory, it is referred to as geometric Jacobian, in contrast with the analytic Jacobian. Furthermore, the action of the Jacobian $ J(X) $ on the joint vector $ \overrightarrow{\xi} $ returns the body velocity $ \eta(X) $ which is expressed in the (micro-)body coordinate frame. Accordingly, $ J(X) $ is called body Jacobian. The relation with the corresponding spatial Jacobian $ ^sJ(X) $, which returns the spatial velocity $ ^s\eta(X) $ expressed in the fixed spatial frame, is obtained by multiplying both side of (\ref{int_vel4}) with $ \mathrm{Ad}_{g(X)} $ and reads:
$$ ^sJ(X)=\mathrm{Ad}_{g(X)}J(X) \; .$$

Finally, by taking the time derivative of (\ref{int_vel4}) the acceleration vector $ \dot{\eta}(X) $ is obtained as:
\begin{equation}\label{int_acc4}
\dot{\eta}(X) = J(X)\ddot{\overrightarrow{\xi}} + \dot{J}(X)\dot{\overrightarrow{\xi}} \; \text{,}
\end{equation}
where $ \dot{J}(X) $ is obtained by a lengthy but straightforward calculation. Defining the $ 6 \times 6 $ components of the Jacobian as $ J(X)= \left[ S_1(X) \;\; S_2(X) \;\; \cdots \;\; S_N(X) \right]$, the time derivative of the Jacobian can be expressed as:
\begin{equation}\label{Jac_dot}
\dot{J}(X) = - \sum_{i=1}^{n-1} \mathrm{ad}_{\sum\limits_{j=i+1}^{n}S_j(X)\dot{\xi}_j}J_i(X) \; \text{,}
\end{equation}
where $ L_{n-1}<X \leqslant L_{n} $ and we have defined $ J_i(X) $ as the Jacobian containing $ 0_6 $ elements except for the $i$-th: $ J_i(X) := [0_6 \cdots S_i(X) \cdots 0_6] \in \mathbb{R}^{6\times6N}$. Alternatively, equation (\ref{int_acc4}) and the expression of Jacobian derivative (\ref{Jac_dot}) can be obtained by back tracking the acceleration $ \dot{\eta}(L_{n-1}) $ and velocity $ \eta(L_{n-1}) $ terms on the right side of (\ref{int_acc3}).

\subsubsection*{Comparison with the PCC Model}
The development above led us to three kinematics equations (\ref{int_pos}), (\ref{int_vel4}) and (\ref{int_acc4}), which give a model to calculate all the kinematic quantities from the knowledge of the joint space of the piece-wise soft arm, in a very similar fashion to traditional rigid manipulators. Compared to the PCC model, the discrete Cosserat approach presented here is able to handle not only constant curvature and elongation, but also shear and torsion, which are fundamental to deal with the strong interactions with the environment characteristic of locomotion and manipulation.

Furthermore, the joint space $ \overrightarrow{\xi} $ composed by the $ N $ constant strains $ \xi_n $ is directly related to the configuration kinematics through the equations (\ref{int_pos}), (\ref{int_vel4}) and (\ref{int_acc4}), while the PCC model needs an additional map between the joint space and the arc parameters space, composed by the length, the curvature and the plane of bending of the section. This allowed us to build a geometric Jacobian instead of an analytic Jacobian, which preserves the natural geometric structure of the motion.

Finally, the intrinsic geometry of the soft robots is reveled. In fact, recognizing (\ref{exp_rototra}) as a screw motion in space, we can conclude that each section forms an arc of screw whose parameters are determined by the constant strain $ \xi_n $ by adapting the formulas normally used for time-twist \cite{Renda_JMR2017}.

\subsection{Piece-wise Constant Strain Dynamics} \label{PWCSD}
In this section we derive the generalized equation of motion of the multi-section piece-wise constant strain model. To that end, we reconsider the continuous dynamics (\ref{dynamics}), that we restate in the weak form of virtual works, {\it i.e.} for any field: $ \delta\zeta(\cdot): X \mapsto \delta\zeta(X)\in \mathfrak{se}(3) $:
\begin{equation}\label{virtual_work}
\int\limits_0^L \delta \zeta^T \left( \mathcal{M} \dot{\eta} + \mathrm{ad}^*_{\eta}\left(\mathcal{M}\eta\right) - \mathcal{F}_{i}'-\mathrm{ad}^*_{\xi}\mathcal{F}_{i} - \bar{\mathcal{F}}_{a} - \bar{\mathcal{F}}_{e} \right) dX = 0
\end{equation}
Note that the above weak form can be derived from the extended Poincar\'e variational calculus of \cite{Boyer_JNLS_2010}. Though being equivalent to the strong form (\ref{dynamics}), this weak form has the advantage of being directly usable to shift the dynamics from the continuous to our piece-wise discrete approach. In fact, to derive the discrete dynamics corresponding to the discrete kinematics (\ref{int_pos2}), it suffices to introduce the relation: $ \delta \zeta(X) = J(X)  \overrightarrow{\delta\xi} $ in addition to the kinematics relations (\ref{int_vel4}) and (\ref{int_acc4}). In these conditions, (\ref{virtual_work}) becomes:
\begin{equation}\label{gen_virtual_work}
\begin{array}{l}
\forall \overrightarrow{\delta\xi} \in \mathfrak{se}(3)^N: \\
\overrightarrow{\delta\xi}^T \int\limits_0^L J^T \left[ \mathcal{M} \left( J \ddot{\overrightarrow{\xi}} + \dot{J}\dot{\overrightarrow{\xi}} \right) + \mathrm{ad}^*_{J\dot{\overrightarrow{\xi}}}\left( \mathcal{M}J\dot{\overrightarrow{\xi}} \right) \right] \\
- J^T \left[ \mathcal{F}_{i}'-\mathrm{ad}^*_{\xi}\mathcal{F}_{i} - \bar{\mathcal{F}}_{a} - \bar{\mathcal{F}}_{e} \right] dX = 0
\end{array}
\end{equation}


which leads to the following generalized dynamics equation once the external loads (\ref{ext_wrench}) and cable driven actuation (\ref{act_wrench}) have been introduced in (\ref{gen_virtual_work}):
\begin{equation}\label{gen_dyn1}
\begin{array}{c}
\left[ \int\limits_0^{L_N}J^T\mathcal{M}_aJ \; dX \right] \ddot{\overrightarrow{\xi}} + \left[ \int\limits_0^{L_N}J^T\mathrm{ad}_{J\dot{\overrightarrow{\xi}}}^*\mathcal{M}_aJ \; dX \right] \dot{\overrightarrow{\xi}} \\
- \left[ \int\limits_0^{L_N}J^T\mathcal{M}_a\dot{J} \; dX \right] \dot{\overrightarrow{\xi}} = - \left[ \int\limits_0^{L_N} J^T \mathcal{D} J \left| J \dot{\overrightarrow{\xi}} \right|_{\mathrm{v}} dX \right] \dot{\overrightarrow{\xi}} \\
 \int\limits_0^{L_N}J^T\left(\mathcal{F}_i' -\mathcal{F}_a' + \mathrm{ad}_{\xi_n}^* \left( \mathcal{F}_i -\mathcal{F}_a \right) \right) \; dX + J(\bar{X})^T\mathcal{F}_p \\
+ \left(1-\rho_w/\rho \right) \left[ \int\limits_0^{L_N}J^T\mathcal{M}\mathrm{Ad}_{g}^{-1} \; dX \right]\mathrm{Ad}_{g_r}^{-1}\mathcal{G} \text{,}
\end{array}
\end{equation}
where, when needed, $ n $ represents the section corresponding to the running value of $ X $ inside the integrals and $ |\cdot|_{\mathrm{v}} $ takes the norm of the translational part of the operand according to equation (\ref{ext_wrench}).

In the remaining part of the section we will describe the different components of (\ref{gen_dyn1}), let us start with the internal elastic and actuation load, those loads are traditionally called
$$ \overrightarrow{\tau} = [\tau_1^T \; \tau_2^T \; \cdots \; \tau_N^T]^T \in \mathbb{R}^{6N} .$$
Due to the linearity of the integral, each element $ \tau_n $ has the form:
$$ \tau_n = \sum_{j=n}^N \int_{L_{j-1}}^{L_j} S_n^T \left( \mathcal{F}_i' -\mathcal{F}_a' + \mathrm{ad}_{\xi_n}^* \left( \mathcal{F}_i -\mathcal{F}_a \right) \right) dX \text{,} $$
where we note that by definition $ S_n(X)=0_6 $ for $ X \leqslant L_{n-1} $ (Fig. \ref{Jacobian_scheme}). Each of the integrals in the series except of the first can be directly solved analytically making use of the identity $ \mathrm{Ad}_g^*\left( \mathcal{F}'+\mathrm{ad}_{\xi}\mathcal{F} \right) = \left( \mathrm{Ad}_{g}^*\mathcal{F} \right)' $, while the first one can be analytically solve with an integration by part with the additional use of the identity $ \mathrm{AD}_{g}^{T\prime} = \mathrm{Ad}_g^{T} = \mathrm{Ad}_{g^{-1}}^{*} $. Applying this operations, we obtain the internal elastic and actuation load for the section $ n $ as follows.
\begin{equation}\label{ela_act_load}
\tau_n = \sum_{j=n}^N \left( S_n^T \left( \mathcal{F}_i-\mathcal{F}_a \right) \right)\rvert_{L_{j-1}}^{L_j} - l\left( \mathcal{F}_{i}-\mathcal{F}_{a} \right) \text{.}
\end{equation}
where $ l $ is the length of the section equal to $ L_n - L_{n-1} $ and we have assumed elastic and actuation loads constant along the section, {\it i.e.}, $ \mathcal{F}_{a}(L_{n-1}<X<L_n) = \mathcal{F}_{a} $ and $ \mathcal{F}_{i}(L_{n-1}<X<L_n) = \mathcal{F}_{i} $ are constants.

In order to calculate the sum in (\ref{ela_act_load}), we exploit the boundary condition at each section. For the cable-driven actuation case they are given below.
\begin{equation}\label{cable_boundary}
\begin{array}{ll}
\mathcal{F}_i\left( L_n^+ \right) = \mathcal{F}_{i(n+1)} & \mathcal{F}_i\left( L_n^- \right) =  \mathcal{F}_{i(n+1)} + \mathcal{F}_{an} \\
\mathcal{F}_a\left( L_n^+ \right) = \sum\limits_{j=n+1}^{N}\mathcal{F}_{aj} & \mathcal{F}_a\left( L_n^- \right) =  \sum\limits_{j=n+1}^N\mathcal{F}_{aj} + \mathcal{F}_{an}
\end{array}
\end{equation}
where the cables are assumed to run from the point of anchorage to the base of the manipulator. The contribution of the cables attached at $ L_n $ is indicated with $ \mathcal{F}_{an} $ and the constant internal load of the section $ n $ with $ \mathcal{F}_{in} $. As expected, crossing an anchoring edge $ L_n $ causes a jump in both the internal elastic and actuation load due respectively to the concentrated load of the cables anchored at that position and the suddenly increase of the number of cable running through the section. Substituting (\ref{cable_boundary}) into (\ref{ela_act_load}), results in a brutal cancellation of the first term (the sum), which becomes:
\begin{equation}\label{cable_act_load}
\tau_n = l \left( \sum_{j=n}^N \mathcal{F}_{aj} - \mathcal{F}_{in} \right)\text{.}
\end{equation}

For what concern the fluidic actuation case, the boundary condition are as follows.
\begin{equation}\label{pneu_boundary}
\begin{array}{ll}
\mathcal{F}_i\left( L_n^+ \right) = \mathcal{F}_{i(n+1)} & \mathcal{F}_i\left( L_n^- \right) =  \mathcal{F}_{i(n+1)} - \mathcal{F}_{a(n+1)} + \mathcal{F}_{an} \\
\mathcal{F}_a\left( L_n^+ \right) = 0_{6 \times 1} & \mathcal{F}_a\left( L_n^- \right) =  0_{6 \times 1}
\end{array}
\end{equation}
where we have taken into account the load exerted at the bottom of the section (Fig. \ref{actuation}) in the jump from $ L_n^+ $ to $ L_n^- $ and the fact that there is no distributed load along the section. Substituting (\ref{pneu_boundary}) into (\ref{ela_act_load}), results in a cancellation of the elastic load in the first term and of the actuation load in the second term, which leads to:
\begin{equation}\label{cable_act_load}
\tau_n = \sum_{j=n}^N \left( S_n^T \left( \mathcal{F}_{aj}-\mathcal{F}_{a(j+1)} \right) \right)\rvert_{L_{j}} - l\mathcal{F}_{in}  \text{.}
\end{equation}
with $ \mathcal{F}_{a(N+1)} = 0_{6 \times 1}$

The second term on the right end side of equation (\ref{gen_dyn1}) represents the generalized external concentrated load and is usually referred to as
$$ \overrightarrow{F} = [F_1^T \; F_2^T \; \cdots \; F_N^T]^T \in \mathbb{R}^{6N} \text{,}$$
where each elements is simply:
\begin{equation}\label{gen_ext_load}
F_{n} = S_n^T(\bar{X})\mathcal{F}_p \text{.}
\end{equation} 

Finally, with those definition at hand and naming the coefficients matrices in squared parenthesis of (\ref{gen_dyn1}), we obtain the piece-wise constant strain dynamic equation:
\begin{equation}\label{gen_dyn2}
\begin{split}
&M\left(\overrightarrow{\xi}\right)\ddot{\overrightarrow{\xi}} + \left( C_1\left(\overrightarrow{\xi},\dot{\overrightarrow{\xi}}\right) - C_2\left(\overrightarrow{\xi},\dot{\overrightarrow{\xi}}\right) \right)\dot{\overrightarrow{\xi}} = \\
&\overrightarrow{\tau}\left(\overrightarrow{\xi}\right) + \overrightarrow{F}\left(\overrightarrow{\xi}\right) + N\left(\overrightarrow{\xi}\right)\mathrm{Ad}_{g_r}^{-1}\mathcal{G} - D\left(\overrightarrow{\xi},\dot{\overrightarrow{\xi}}\right) \dot{\overrightarrow{\xi}} \; \text{,}
\end{split}
\end{equation}
where we recognize the structure of the Lagrangian model of rigid serial manipulators.

Let us now break down each matrix coefficients of the dynamic equation (\ref{gen_dyn2}). Looking at (\ref{gen_dyn1}), the mass matrix $ M\left(\overrightarrow{\xi}\right) \in \mathbb{R}^{6N \times 6N} $ is a symmetric, positive define matrix and his $ 6 \times 6 $ block-element of block-row $ n $ and block-column $ m $ is calculated as follows.
\begin{equation}\label{mass_matrix}
M_{(n,m)} = \sum_{i=max(n,m)}^N\int_{L_i-1}^{L_i}S_n^T\mathcal{M}_a S_m \; dX \text{,}
\end{equation}
where we have exploited the fact that for $ i < max(n,m) $ or equivalently $ X < L_{max(n,m)-1} $ either $ S_n(X) $ or $ S_m(X) $ is equal to $ 0_6 $ (Fig. \ref{Jacobian_scheme}). Similarly, for the Coriolis matrices $ C_1\left(\overrightarrow{\xi},\dot{\overrightarrow{\xi}}\right) \text{, } C_2\left(\overrightarrow{\xi},\dot{\overrightarrow{\xi}}\right) \in \mathbb{R}^{6N \times 6N} $ we obtain
\begin{equation}\label{coriolis1_matrix}
C_{1(n,m)} = \sum_{i=max(n,m)}^N\int_{L_i-1}^{L_i}S_n^T\mathrm{ad}_{J\dot{\overrightarrow{\xi}}}^*\mathcal{M}_aS_m \; dX \text{,}
\end{equation}
\begin{equation}\label{coriolis2_matrix}
C_{2(n,m)} = \sum_{i=max(n,m)}^N\int_{L_i-1}^{L_i}S_n^T\mathcal{M}_a\mathrm{ad}_{\sum\limits_{j=m+1}^{i}S_j\dot{\xi}_j}S_m \; dX \text{,}
\end{equation}
while, for the drag matrix $ D\left(\overrightarrow{\xi},\dot{\overrightarrow{\xi}}\right) \in \mathbb{R}^{6N \times 6N} $, we get:
\begin{equation}\label{drag_matrix}
D_{(n,m)} = \sum_{i=max(n,m)}^N\int_{L_i-1}^{L_i}S_n^T\mathcal{D}S_m\left| J \dot{\overrightarrow{\xi}} \right|_{\mathrm{v}} \; dX \text{.}
\end{equation}
With the same reasoning, the block-element of block-row $ n $ of the gravitational-buoyancy matrix $ N\left(\overrightarrow{\xi}\right) \in \mathbb{R}^{6N \times 6} $ is as follows
\begin{equation}\label{grav_matrix}
N_{(n)} = \left(1-\rho_w/\rho \right) \sum_{i=n}^N\int_{L_i-1}^{L_i}S_n^T\mathcal{M}\mathrm{Ad}_{g}^{-1} \; dX \text{.}
\end{equation}

\begin{figure*}[tb]
\centering
\captionsetup{justification=centering}
\begin{equation*}
\begin{array}{cc}
\begin{array}{ccc}
			   & \xrightarrow{\makebox[6.5cm]{$X$}} 	&   \\
J(X)^T =  & \left\{\begin{array}{ccccc} \left[ \begin{array}{c} S_1^T \\ 0_6 \\ 0_6 \\ 0_6 \\ \vdots \\ 0_6 \end{array} \right] & \left[ \begin{array}{c} S_1^T \\ S_2^T \\ 0_6 \\ 0_6 \\ \vdots \\ 0_6 \end{array} \right] & \left[ \begin{array}{c} S_1^T \\ S_2^T \\ S_3^T \\ 0_6 \\ \vdots \\ 0_6 \end{array} \right] & \cdots & \left[ \begin{array}{c} S_1^T \\ S_2^T \\ S_3^T \\ S_4^T \\ \vdots \\ S_N^T \end{array} \right] \end{array}\right. & \xdownarrow{1.5cm}{n}
\end{array} &
\begin{array}{ccc}
			   & \xrightarrow{\makebox[2.7cm]{$n$}} 	&   \\
J(X) =  & \left\{\begin{array}{c} \left[ S_1 \; 0_6 \; 0_6 \; 0_6 \; \cdots \; 0_6 \; \right] \\ \\
						  		  \left[ S_1 \; S_2 \; 0_6 \; 0_6 \; \cdots \; 0_6 \; \right] \\ \\
						  		  \left[ S_1 \; S_2 \; S_3 \; 0_6 \; \cdots \; 0_6 \; \right] \\ \\
						  		  \vdots 											   \\ \\	 		  							  \left[ S_1 \; S_2 \; S_3 \; S_4 \; \cdots \; S_N \; \right] \\ \\
				 \end{array}\right. & \xdownarrow{2cm}{X}
\end{array}
\end{array}
\end{equation*}
\caption{Schematic of the Jacobian and its transposed highlighting the structure with respect to the position $ X $ and section $ n $.}
\label{Jacobian_scheme}
\end{figure*}

We have now all the ingredients to process the joint dynamic (\ref{gen_dyn2}) and reconstruct the shape, velocity and acceleration of the soft manipulator with (\ref{int_pos}), (\ref{int_vel4}) and (\ref{int_acc4}).

\section{Simulation Results}\label{SR}

In this section the PCS dynamic model (\ref{gen_dyn2}) is tested through different simulations. First, a plane motion of three sections manipulator is shown, then an out-of-plane motion involving the torsion of all the three sections, which is not possible with the PCC model, is performed. Finally, in order to show how the PCS model copes with non constant external load, the model is compared with a cantilever beam, simulated using the continuous Cosserat model (\ref{dynamics}). Before that, an efficient recursive algorithm aiming to calculate the coefficient matrices of (\ref{gen_dyn2}) is presented. 

\subsection{Recursive Algorithm}
\begin{figure*}
\centering
\includegraphics[scale=0.28]{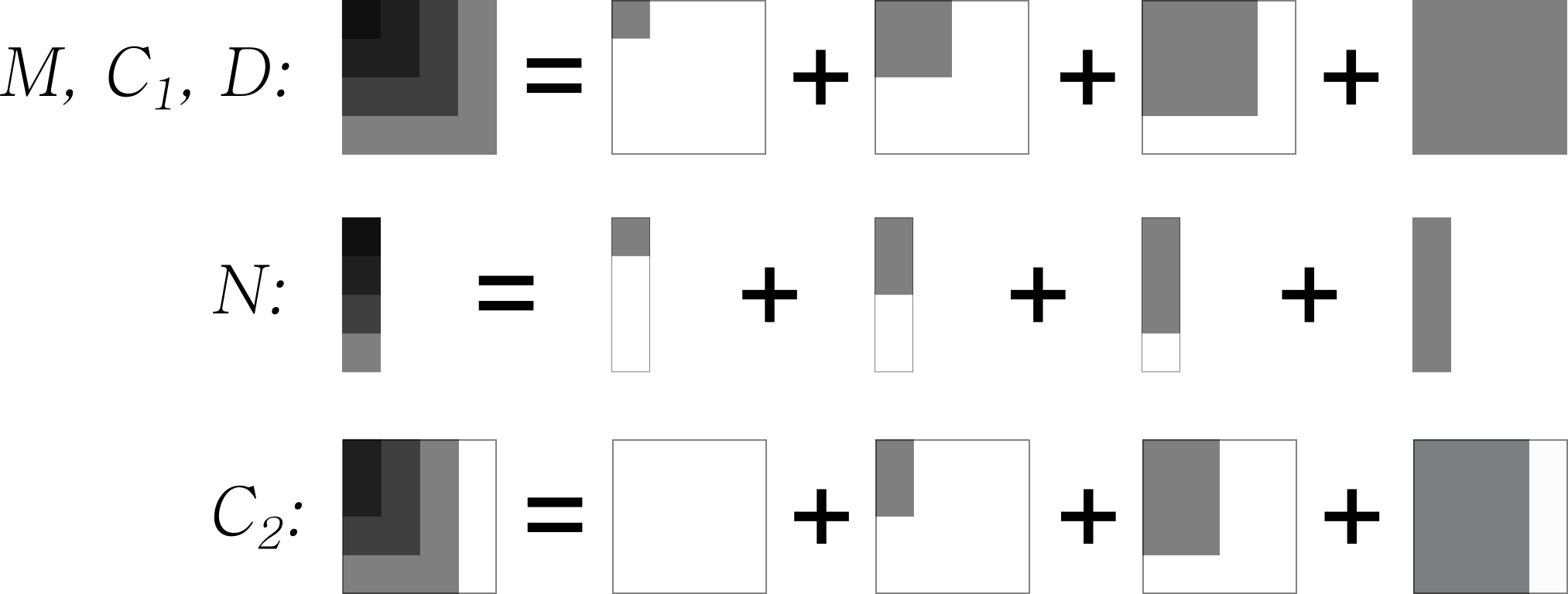}
\caption{Composition of the coefficient matrices from the contribution of the non zeros block-matrices due to each section ({\it e.g.} four sections).}
\label{recursivity}
\end{figure*}

The basic idea for the recursive algorithm is that each section of the soft manipulator contributes to a very specific set of block-elements of the coefficient matrices. In particular, the non zero block-elements due to section $ n $ of the mass matrix $ M $, the first Coriolis matrix $ C_1 $ and the drag matrix $ D $ are those located in the square block-matrix of block-rows 1 to $ n $ and block-column 1 to $ n $, while for the gravitational matrix $ N $, they compose the block-rows from 1 to $ n $ of the only block-column and finally, for the second Coriolis matrix $ C_2 $, the non zero block-elements form a rectangular block-matrix of block-rows 1 to $ n $ and block-column 1 to $ n-1 $ (Figure \ref{recursivity}). This can be seen by splitting the integrals in the coefficient matrices of equation (\ref{gen_dyn1}) into the $ N $ integrals corresponding to each section, then the non zero square block-matrix rise from the  varying structure of the Jacobian $ J(X) $ (and the modified Jacobian for $ C_2 $) with respect to $ X $ as shown in figure \ref{Jacobian_scheme}.

An efficient way to implement this technique, is to benefit from the results of the calculations given by the previous section. Going further into details, at a certain $ L_{n-1} < X \leqslant L_n $ the Jacobian element $ S_{n-1}(X) $ has only one member which actually depends on $ X $, all the rest being inherited from the last evaluation of the same quantity in the previous section $\left( S_{n-1}(L_{n-1}) \right) $, as it can be visualized by inspecting equation (\ref{Jacobian}). This is used in the calculation of $ J(X) $ and the adjoint elements of $ C_2 $ (after multiplication with $ \dot{\xi}_{n-1} $). Furthermore, $ \eta(L_{n-1}) $ and $ g(L_{n-1}) $ are calculated respectively through equation (\ref{int_vel2}) and (\ref{int_pos2}) to obtain the co-adjoint member in $ C_1 $ and the Adjoint member in $ N $.

\subsection{Plane \& Out-of-Plane Motion}
In this section the feasibility of the model to perform highly dynamic motion both in plane and out-of-plane is shown. The simulated soft manipulator is composed by three cylindrical sections of length $ l = 250 $ $ mm $, radius equal to $ 10 $ $ mm $, Young modulus $ E = 110 $ $ kPa $, shear viscosity modulus $ \upsilon = 0.3 $ $ kPasec $, Poisson modulus equal to $ 0.5 $ and mass density $ \rho = 1080 $ $ kg/m^3 $. The manipulator lays upside-down as shown in figure \ref{kinematics} and shares the $ Z $ axis with the inertial frame, therefore the transformation map between the spatial frame and the base frame is: $ g_r = diag(-1, -1, 1, 1) $. The actuation load in both cases is fluidic and imposed over time through a ramp starting from zero, with a 1 second width, and reaching in the planar case:
$$ \mathcal{F}_{a1} = \left( \begin{array}{c} 0 \\ 0 \\ 10 \\ 0 \\ 0 \\ 0  \end{array} \right) \text{ , } \mathcal{F}_{a2} = \left( \begin{array}{c} 0 \\ 0 \\ -4 \\ 0 \\ 0 \\ 0  \end{array} \right) \text{ , } \mathcal{F}_{a3} = \left( \begin{array}{c} 0 \\ 0 \\ 2 \\ 0 \\ 0 \\ 0 \end{array} \right) \text{ , } $$
and in the out-of-plane case:
$$ \mathcal{F}_{a1} = \left( \begin{array}{c} -0.5 \\ 0 \\ 5 \\ 0 \\ 0 \\ 0 \end{array} \right) \text{ , } \mathcal{F}_{a2} = \left( \begin{array}{c} -0.5 \\ 1.25 \\ 0 \\ 0 \\ 0 \\ 0 \end{array} \right) \text{ , } \mathcal{F}_{a3} = \left( \begin{array}{c} -0.5 \\ 0 \\ -0.5 \\ 0 \\ 0 \\ 0 \end{array} \right) \text{ , } $$
where all the value are in $ 10^{-3}Nm $. The gravity load has been neglected.

\begin{figure}
\centering
\includegraphics[scale=0.50]{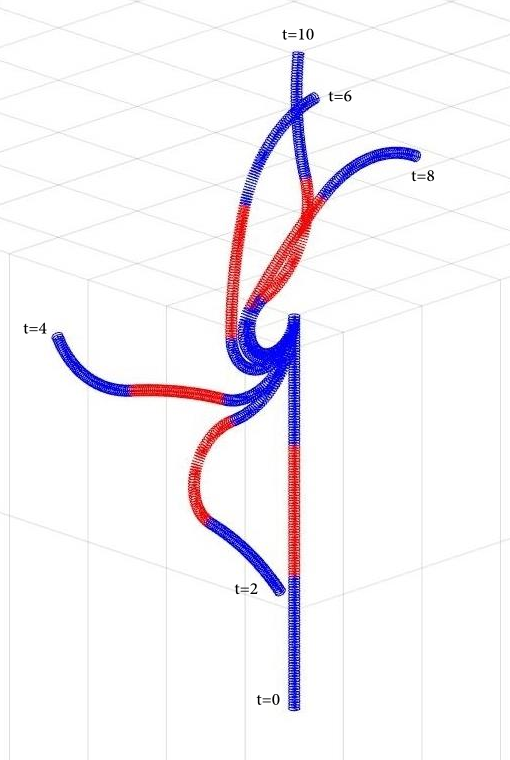}
\caption{Plane motion snapshots at time $ t = 0 $, $ 2 $, $ 4 $, $ 6 $, $ 8 $ and $ 10 $.}
\label{plane_motion}
\end{figure}

\begin{figure}
\centering
\includegraphics[scale=0.4]{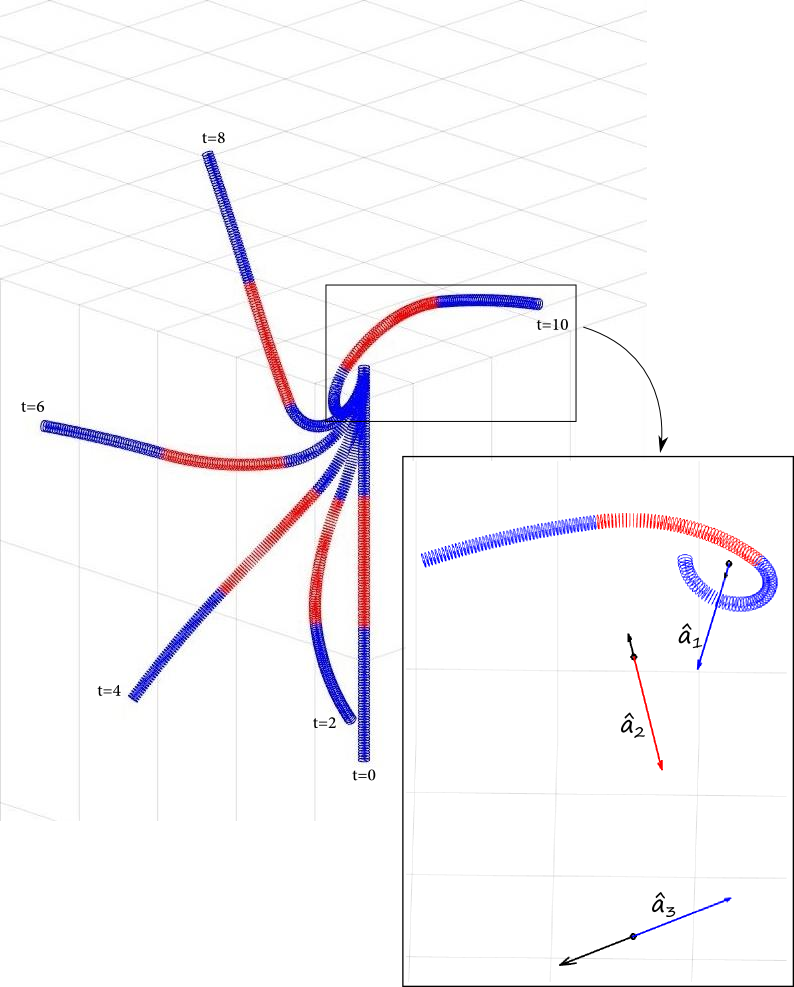}
\caption{Out-of-plane motion snapshots at time $ t = 0 $, $ 2 $, $ 4 $, $ 6 $, $ 8 $ and $ 10 $. For the last configuration, highlighted in the square box, the colored arrows represent the axis of the three screws $ \hat{a}_n $ and the black arrows indicate the amount of translation in the direction of the screw.}
\label{outplane_motion}
\end{figure}

Few snapshots of the plane motion are shown in figure \ref{plane_motion}, while the out-of-plane motion is shown in figure \ref{outplane_motion}. The screws associated with the last configuration of the out-of-plane motion are also shown in figure \ref{outplane_motion}. Using the terminology of screw \cite{Murray}, the colored arrows represent the axis $ \hat{a}_n $ of the three screws, around which the sections rotate of an amount equal to the magnitude $ m_n $, while the black arrows indicate the amount of translation in the direction of the screw, given by $ h_n m_n $, $ h_n $ being the pitch of the screw. For this particular configuration we obtained:
$$ \begin{array}{ccc}
h_1 = 8 \; mm & h_2 = -45 \; mm & h_3 = -683 \; mm \\
m_1 = 3.9 & m_2 = 1  & m_3 = 0.3  \text{.}
\end{array} $$

\subsection{Cantilever Beam Comparison}
Even if one could manage to design the actuation of a soft manipulator in order to be constant in each section, non-constant loads due to gravity, external forces and inertial forces are unavoidable in realistic condition. For this reason, it is important to know how the discrete Cosserat model handles a non-constant load and what are the effects of such loads in terms of accuracy of the result. To do so, we notice that each member of the dynamic equation (\ref{gen_dyn2}) is pre-moltiplied by $ J^T(X) $ or, block-element-wise, by $ S_n^T(X) $. Now, looking at the Jacobian (\ref{Jacobian}), we see that an element $ S_n^T(X) $ will first map the considered load to the base of the section $ n $ through $ \mathrm{Ad}^* $ and then integrate all the "re-maps" of this load up to $ X $ through
$$ \mathrm{AD}_{g_n}^T(X) = \int_{L_{n-1}}^{X}\mathrm{Ad}_{g_n(s)}^T ds = \int_{L_{n-1}}^{X}\mathrm{Ad}_{g_n^{-1}(s)}^* ds $$
which gives, by definition of integral, $ l $ times the mean of this load on the section $ n $. It is worth to highlight here, that this is in essence how the discerete Cosserat model relates the ideal assumption of piece-wise constant strains with the real continuously varying counterpart.

Intuitively, the wider is the interval in which the mean is evaluated the larger is the discrepancy with the real distribution. In order to show this fact and test the model with a non-constant load scenario, the continuous Cosserat model (\ref{dynamics}) and the discrete Cosserat model (\ref{gen_dyn2}) are applied in the following to a cantilever beam with vertical tip load. The simulated beam is of cilindrical shape, with length $ L = 250 $ $ mm $, radius equal to $ 10 $ $ mm $, Young modulus $ E = 110 $ $ kPa $, shear viscosity modulus $ \upsilon = 0.3 $ $ kPasec $, Poisson modulus equal to $ 0 $ and mass density $ \rho = 2000 $ $ kg/m^3 $. The beam lays on the right side  of the inertial frame (($ e_1, e_2, e_3 )$ in figure \ref{kinematics}) and share the $ Z $ axis with this frame, therefore the map between the spatial frame and the base frame is:
$$ g_r = \left( \begin{array}{cccc} 0 & -1 & 0 & 0 \\ 1 & 0 & 0 & 0 \\ 0 & 0 & 1 & 0 \\ 0 & 0 & 0 & 1 \end{array} \right) \text{.} $$
Finally, the external tip load points in the positive $ y $ direction with respect to the fixed base frame and is applied at $ \bar{X} = L $, thus it has the form:
$$ \mathcal{F}_p = \left( \begin{array}{cc} R^T(L) & 0 \\ 0 & R^T(L) \end{array} \right) \left( \begin{array}{c} 0 \\ 0 \\ 0 \\ 0 \\ 10 \\ 0 \end{array} \right) \text{,} $$
with unit reference of $ 10^{-3}N $.

\begin{figure}
\centering
\includegraphics[scale=0.50]{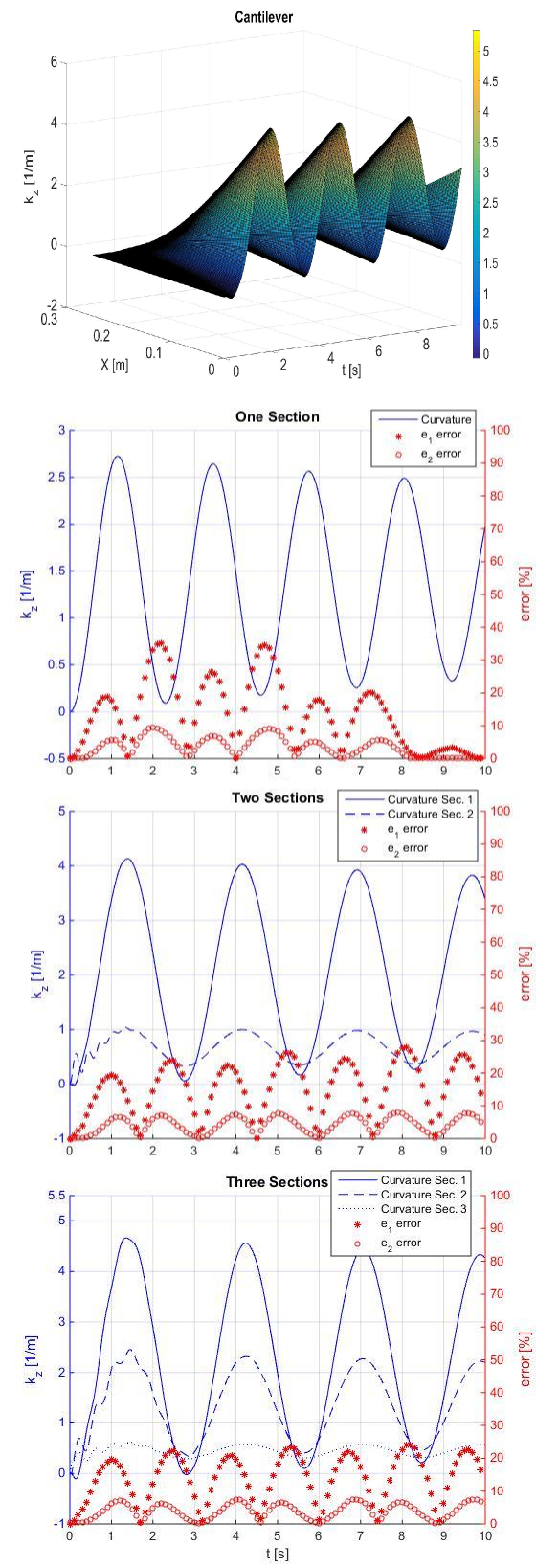}
\caption{From top, curvature of the continuous cantilever as a function of space and time, followed by the curvatures of the discrete cantilever divided in one, two and three sections (blue lines) along with the tip position error expressed as percentage of the total length $ L $ (red markers, in both the $ e_1 $ and $ e_2 $ directions).}
\label{cantilever_comparison}
\end{figure}

In figure \ref{cantilever_comparison}, on top, is shown the resulting curvature of the continuous cantilever as a function of space and time, followed by the curvatures of the discrete cantilever divided in one, two and three sections (blue lines). We immediately notice that the oscillation frequency for the one section case is much higher than that of the continuous cantilever. This discrepancy is quantified at each time by the tip position error expressed in percentage of the total length $ L $ (red markers, in both the $ e_1 $ and $ e_2 $ directions). Intuitively, this can be explained by the fact that the additional constraint of  constant strains applied to the Cosserat micro-solids in the discrete model makes the system more rigid. As expected, the oscillation frequency progressively slows down toward the continuous value with the increase of the number of sections while, accordingly, the error gradually decreases.

The additional rigidity due to the constant strain constraint is confirmed by the steady state comparison. As a matter of fact, in all the three cases with one, two and three sections, the steady state tip position is above the real cantilever tip position, or in other words the beam is less deformed. Again, the error decreases in both the $ e_1 $ and $ e_2 $ directions with the increase of sections. The respective steady-state error values are shown below.
\begin{equation*}
\begin{array}{cccc}
	&  \text{one section} &  \text{two sections} &  \text{three sections} \\
\text{error [\%] } e_1 & 5.55  & 1.58 & 1.42 \\
\text{error [\%] } e_2 & 1.29  & 0.48 & 0.07
\end{array}
\end{equation*}

\section{Experimental Results}\label{ER}

In this section the PCS dynamic model performances are compared against experimental data. In order to evaluate the results with respect to the continuous Cosserat model, we have used the same prototype, parameters and experimental data provided in \cite{Renda_TRO}, which we refer to for more  exhaustive details on the experimental platform and measurement set up.

\begin{figure}
\centering
\includegraphics[scale=0.40]{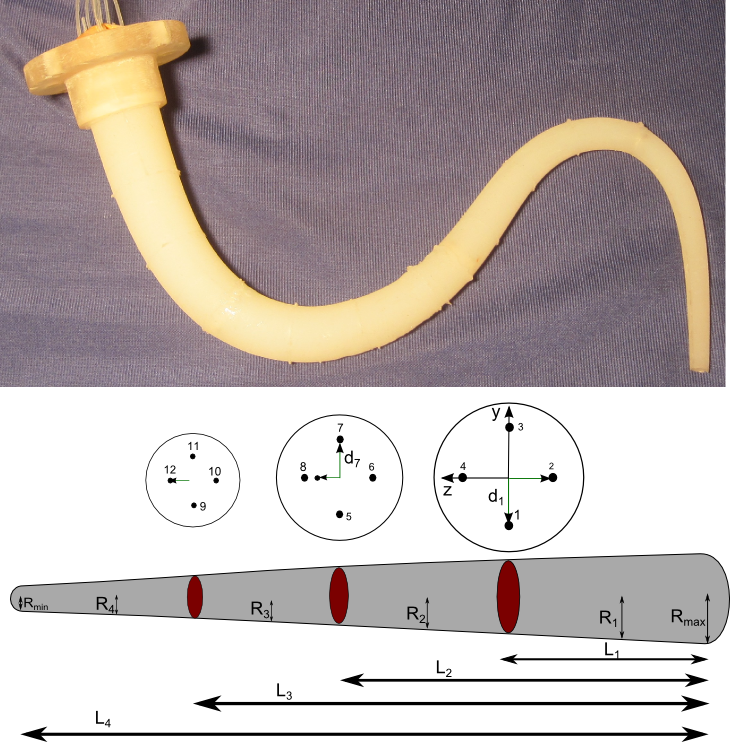}
\caption{Real prototype (top) and schematic (bottom) of the soft manipulator used in the experiments.}
\label{prototype_design}
\end{figure}

In short, the prototype is composed of a single conical piece of silicone, with a base radius $ R_{max} $ and a tip radius $ R_{min} $, actuated by 12 cables embedded inside the robot body. The cables run parallel to the midline at a distance $ d_j $ ($ j \in \{1,2, \dots , 12\} $) and are anchored four at a time at three different lengths along the robot arm ($ L_1 $, $ L_2 $, $ L_3 $) and with a relative angle of 90 degrees (Figure \ref{prototype_design}). During operation, the cable tensions, driven by servomotors, are measured by force sensors while the motion of the arm is recorded with two high speed cameras. The 3-D motion is then reconstructed through a process based on the direct linear transformation (DLT).

The soft manipulator has been tested for three different conditions, a single bending motion produced by cable 11, an in-plane multi-bending produced by a sequence of activation of cables 9, 11, 1, 3 and an out-of-plane multi-bending produce by cables 11, 5, 2. The details of the cable activaion is reported in Figure \ref{experimental_res} (top three graphs).

In \cite{Renda_TRO}, it has been found that the drag and added mass matrix in this case can be expressed as
$$ \mathcal{D} = \left( \begin{array}{cc} 0 & 0 \\ 0 & \mathsf{D} \end{array} \right) \text{,} \;\;\;\; 
\mathcal{A} = \left( \begin{array}{cc} 0 & 0 \\ 0 & \mathsf{F} \end{array} \right) \text{,} $$
where $ \mathsf{D}(X) = diag(1/2 \pi R C_x, R C_y, R C_z)\rho_w $ and $ \mathsf{F}(X) = diag(0, A B_y, A B_z)\rho_w $, $ R(X) $ being the radius of the soft arm and $ C_x $, $ C_y $, $ C_z $, $ B_y $, $ B_z $ being fluid dynamics coefficients. The mechanical and geometrical parameters of the arm are summarized in Table \ref{parameters}.

\begin{table}
\normalsize
\caption{Design Parameters of the Prototype}
\label{parameters}
\centering
\begin{tabular}{|c|c||c|c|}
\hline Parameter & Value & Parameter & Value\\ 
\hline $ R_{max} $ & $ 15 $ $ mm $ & $ d_1 $, $ d_2 $, $ d_3 $, $ d_4 $ & $ 9 $ $ mm $\\ 
\hline $ R_{min} $ & $ 4 $ $ mm $ & $ d_5 $, $ d_6 $, $ d_7 $, $ d_8 $ & $ 6 $ $ mm $\\ 
\hline $ L_{1} $ & $ 98 $ $ mm $ & $ d_9 $, $ d_{10} $, $ d_{11} $, $ d_{12} $ & $ 3 $ $ mm $ \\ 
\hline $ L_{2} $ & $ 203 $ $ mm $ & $ gr $ & $ 9.81 $ $ \frac{m}{s^2} $\\ 
\hline $ L_{3} $ & $ 311 $ $ mm $ & $ C_x $ & $ 0.01 $ \\
\hline $ L_{4} $ & $ 418 $ $ mm $ & $ \rho_{w} $ & $ 1.022 $ $ \frac{kg}{dm^3} $\\ 
\hline $ E $ & $ 110 $ $ kPa $ & $ C_y $ & $ 2.5 $\\
\hline $ \mu $ & $ 300 $ $ Pa \cdot s $ & $ C_z $ & $ 2.5 $\\
\hline $ \nu $ & $ 0.5 $ & $ B_y $ & $ 1.5 $\\ 
\hline $ \rho $ & $ 1.08 $ $ \frac{kg}{dm^3} $ & $ B_z $ & $ 1.5 $\\ 
\hline
\end{tabular}
\end{table}

\subsection{Comparison}
\begin{figure*}
\centering
\includegraphics[scale=0.55]{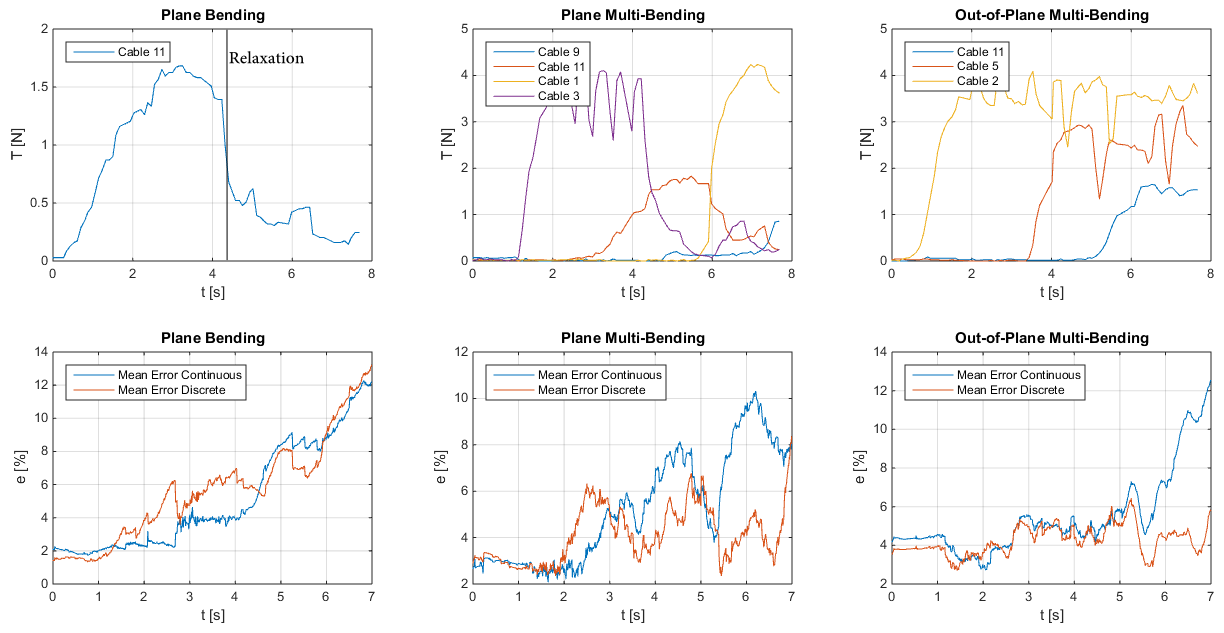}
\caption{Measured cable tensions which represent the input of the model (top) and experimental results (bottom) with a comparison between the results of the continuous model of \cite{Renda_TRO} and the present discrete model.}
\label{experimental_res}
\end{figure*}

In order to exploit the dynamics equations developed in \ref{PWCSD}, the soft manipulator has been modeled as a stack of four cilindrical constant-strain sections defined by $ L_1 $, $ L_2 $, $ L_3 $, $ L_4 $, with a radius equal to the mean of the prototype radius for each section ($ R_1 $, $ R_2 $, $ R_3 $ and $ R_4 $ in Figure \ref{prototype_design}). The dynamics and kinematics equations have been solved  by implementing the recursive algorithm presented above for the three load conditions of the experiments. The results of the tests are reported in Figure \ref{experimental_res} together with the one obtained in \cite{Renda_TRO} with the continuous Cosserat model. The error is calculated as the normalized mean at each time step of the Euclidean distance between the simulated and real markers positioned a the tip of each section.

\subsection{Discussion}
As can be observed from Figure \ref{experimental_res}, the results of the present discrete model are comparable or even better than the ones obtained with the continuous model, as a matter of fact, the average error in the three cases are respectively 5.1\%, 5.2\% and 5.4\% for the continuous model versus 5.7\%, 4.2\% and 4.3\% for the discrete model. In the authors opinion, the reason for this improvement lays mainly on the higher numerical stability shown by the PCS discrete model, in particular, on the different management of the internal point load exerted where the cables are fastened. In the continuous model an internal point load is modeled with a Dirac function that has to be discretized during the numerical integration, while in the PCS model there is no such approximation and the concentrated load is introduce naturally with the boundary conditions for each section.

In order to further improve the accuracy of the model, the friction of the cables should be included which in turn models the hysteresis behavior of the load-unload cycle. As it is highlighted in \cite{Renda_TRO}, the hysteresis behavior is clear from the plane bending experiments, in which the error increase drastically after the relaxation of the cable which is when the load is mainly driven by the friction of the cable against the silicone body. From a geometric point of view, the number of discrete sections could be increased in order to capture non-negligible variation of the strain due to external loads. Furthermore, the model could be able to better take into account the variation with respect to $ X $ of the mass $ \mathcal{M}(X) $ in the calculus of the mass matrix $ M $ as well as the variation of the stiffness and viscosity matrices $ \Sigma(X) $, $ \Upsilon(X) $ in the calculus of the internal elastic load of $ \overrightarrow{\tau} $, which are due to the conical shape of the manipulator.

\section{Conclusion}

In conclusion, a new piece-wise constant strain model for multi-section soft robots has been presented which is based on the discretization of the continuous Cosserat model inheriting from it the fruitful geometrical and mechanical properties. The close relation between this model for soft robotics and the traditional model for rigid robotics is also highlighted. The PCS model has been extensively corroborated through simulation and experimental results of plane and out-of-plane multi-bending. Furthermore, the performances have been compared with the continuous Cosserat model showing comparable or even better results. It is worth to highlight, that a similar approach can be found in the context of recent finite element formulation for geometrically exact beam as in \cite{Sonneville_CMAME2014}, making of this work a bridge between different engineering disciplines.

\bibliographystyle{plain}
\bibliography{Renda_BIB}

\begin{IEEEbiography}[{\includegraphics[scale=0.80]{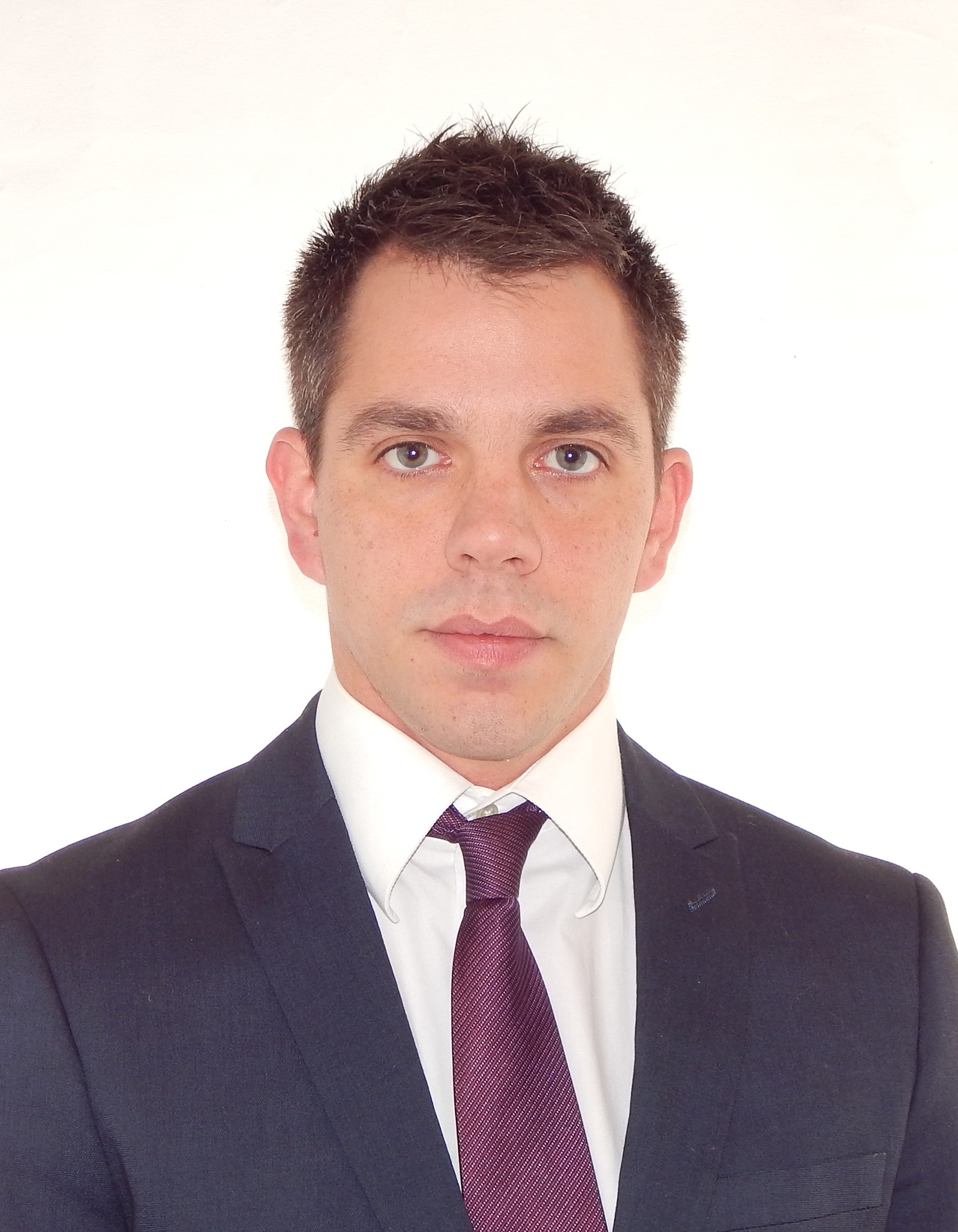}}]{Dr. Federico Renda}
received his BEng and MEng degrees in Biomedical Engineering in 2007 and 2009 respectively from the University of Pisa. He completed his PhD in Bio Robotics in 2014 from the BioRobotics Institute of the Scuola Superiore SantAnna, Pisa. In 2013, as visiting PhD student, he joined the IRCCyN Lab, at the Ecole des Mines de Nantes, Nantes. He is currently a Post Doctoral Fellow with the Khalifa University Robotics Institute of Khalifa University of Science, Technology and Research, Abu Dhabi. His research activity include the geometrically exact modeling of soft robots and its application in design and control.
\end{IEEEbiography}

\begin{IEEEbiography}[{\includegraphics[scale=0.80]{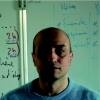}}]{Prof. Fr\'ed\'eric Boyer}
was born in France in 1967. He received both the Diploma degree in mechanical engineering and the Master of Research degree in mechanics from the Institut Nationale Polytechnique de Grenoble, Grenoble, France, in 1991, and the Ph.D. degree in robotics from the University of Paris VI,
Paris, France, in 1994. He is currently a Professor with the Department of Automatic Control, Ecole des Mines de Nantes, Nantes, France, where he works with the Robotics Team, Institut de Recherche en Communication et Cybern\'etique de Nantes. His current research interests include structural dynamics, geometric mechanics, and bio-robotics. Prof. Boyer received the Monpetit Prize from the Academy of Science of Paris in 2007 for his work in dynamics.
\end{IEEEbiography}

\begin{IEEEbiography}[{\includegraphics[scale=0.50]{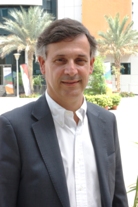}}]{Prof. Jorge Dias}
has a Ph.D. on Electrical Engineering by the University of Coimbra, Portugal, specialization in Control and Instrumentation. Jorge Dias have been professor at the Department of Electrical Engineering and Computers and researcher from the Institute of Systems and Robotics (ISR) from the University of Coimbra (UC). Jorge Dias research is in the area of Computer Vision and Robotics and has contributions on the field since 1984. He has several publications in international journals, books, and conferences. Jorge Dias coordinates the research group for Artificial Perception for Intelligent Systems and Robotics of Institute of Systems and Robotics from University of Coimbra and the Laboratory of Systems and Automation (http://las.ipn.pt) from the Instituto Pedro Nunes (IPN) a technology transfer institute from the University of Coimbra, Portugal. Since July 2011, Jorge Dias is on leave of absence to setup the Robotics Institute and research activities on robotics at Khalifa University (Abu Dhabi, UAE).
\end{IEEEbiography}

\begin{IEEEbiography}[{\includegraphics[scale=0.50]{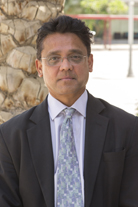}}]{Prof. Lakmal Seneviratne}
obtained a BSc (Eng.) and PhD in Mechanical Engineering from King’s College London. He was a research associate at University College London and an engineer at GEC Energy Systems Ltd, prior to joining King’s College London as a lecturer. He is currently a Professor of Mechatronics. His research interests include robotics and automation, with special emphasis on the use of engineering mechanics based algorithms to create intelligent behaviour in a variety of mechatronic applications.
\end{IEEEbiography}

\end{document}